\documentclass[letterpaper, 10 pt, conference]{ieeeconf}  

\IEEEoverridecommandlockouts                              

\usepackage{amsmath} 
\usepackage{amssymb}  
\usepackage{mathtools}
\usepackage[acronym]{glossaries}

\DeclarePairedDelimiterX{\infdivx}[2]{(}{)}{%
  #1\;\delimsize\|\;#2%
}
\newcommand{\kldiv}{D_{KL}\infdivx}

\newacronym{mis}{MIS}{minimally-invasive surgery}
\newacronym{rl}{RL}{reinforcement learning}
\newacronym{sac}{SAC}{soft actor critic}
\newacronym{ppo}{PPO}{proximal policy optimization}
\newacronym{trpo}{TRPO}{trust region policy optimization}
\newacronym{pi2}{PI$^2$}{policy improvement with path integrals}
\newacronym{her}{HER}{hindsight experience replay}
\newacronym{vr}{VR}{virtual reality}
\newacronym{fem}{FEM}{finite element method}
\newacronym{fps}{FPS}{frames per second}
\newacronym{fft}{FFT}{fast Fourier transform}

\title{\LARGE \bf
RL-Based Guidance in Outpatient Hysteroscopy Training: A Feasibility Study
}

\author{Vladimir Poliakov$^{1}$ $^{2}$, Kenan Niu$^{2}$, Emmanuel Vander Poorten$^{2}$, and  Dzmitry Tsetserukou$^{1}$
//
\thanks{$^{1}$ Vladimir Poliakov and Dzmitry Tsetserukou are with the Intelligent Space Robotics Lab, Skolkovo Institute of Science and Technology, 121205 Moscow, Russia}%
\thanks{$^{2}$ Vladimir Poliakov, Kenan Niu and Emmanuel Vander Poorten are with the Robot Assisted Surgery group, Faculty of Mechanical Engineering, KU Leuven, 3001 Leuven, Belgium
        {\tt\small vladimir.poliakov@kuleuven.be}}%
}

\begin{document}

\maketitle
\thispagestyle{empty}
\pagestyle{empty}

\begin{abstract}

This work presents an RL-based agent for outpatient hysteroscopy training. Hysteroscopy is a gynecological procedure for examination of the uterine cavity. Recent advancements enabled performing this type of intervention in the outpatient setup without anaesthesia. While being beneficial to the patient, this approach introduces new challenges for clinicians, who should take additional measures to maintain the level of patient comfort and prevent tissue damage. Our prior work has presented a platform for hysteroscopic training with the focus on the passage of the cervical canal. With this work, we aim to extend the functionality of the platform by designing a subsystem that autonomously performs the task of the passage of the cervical canal. This feature can later be used as a virtual instructor to provide educational cues for trainees and assess their performance. The developed algorithm is based on the soft actor critic approach to smooth the learning curve of the agent and ensure uniform exploration of the workspace. The designed algorithm was tested against the performance of five clinicians. Overall, the algorithm demonstrated high efficiency and reliability, succeeding in 98\% of trials and outperforming the expert group in three out of four measured metrics.
\end{abstract}

\section{INTRODUCTION}
\begin{figure}[ht]
    \centering
    \includegraphics[width=0.8\linewidth]{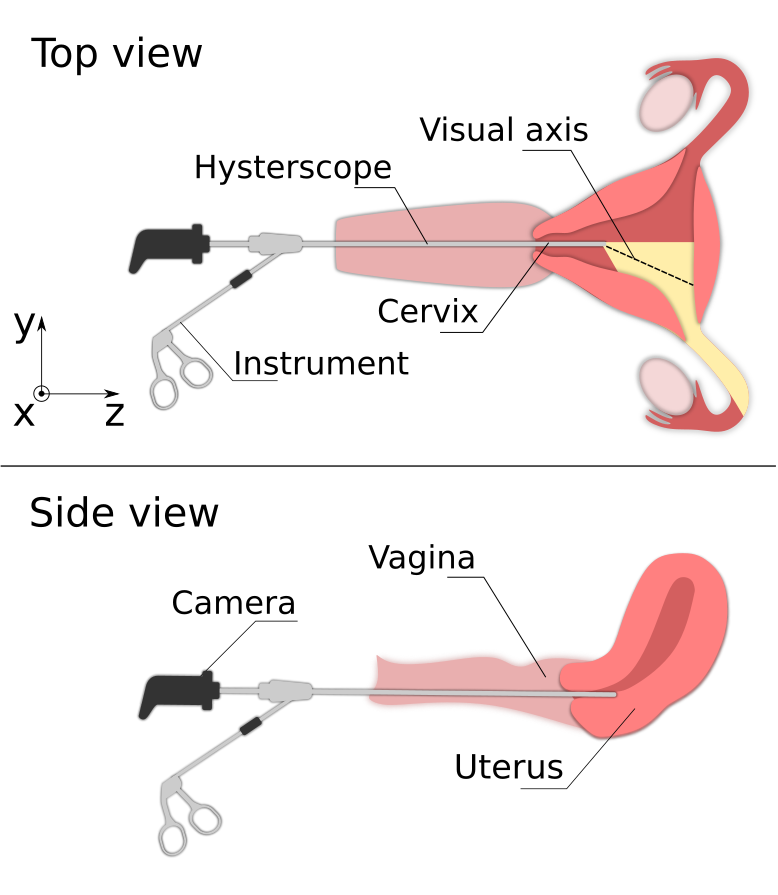}
    \caption{The schematic diagram shows top and side views of the hysteroscopic procedure. The hysteroscope is inserted through the vaginal canal
    and the endocervix to inspect and treat the uterine cavity.}
    \label{fig:hysteroscopy}
\end{figure}

Hysteroscopy is the type of gynecological procedure for diagnosis and treatment of intrauterine pathology by the means of  \gls{mis}. Figure \ref{fig:hysteroscopy} depicts the schematic representation of the procedure. A long slender telescope, the hysteroscope, is inserted through the vaginal canal and then advanced through the cervical canal to access the uterine cavity without additional incisions.  The hysteroscope is typically a straight rigid tube with the outer diameter of approximately five millimeters. The hysteroscope is equipped with a working channel to deploy hysteroscopic instruments, e.g. forceps or scissors, with which a clinician can perform manipulations on tissue. 
Modern techniques allow to perform hysteroscopy in the outpatient setup without anaesthesia, this way reducing the recovery time and mitigating the risks associated with preparatory medication \cite{BETTOCCHI20091}. This type of hysteroscopy is typically referred to as in-office hysteroscopy. 

A broad variety of uterine abnormalities can be diagnosed and treated on the spot using in-office hysteroscopy without major risks to the patient \cite{vanWesselSteffi2018HitN}. However, some parts of the procedure still remain to be challenging and require high level of concentration. In particular, the passage of the cervix is considered to be the bottle neck of hysteroscopy \cite{Hernandez2018}. The cervical canal is a narrow and curved orifice that provides an entry point into the uterine cavity. The diameter of a normal cervical canal is typically within the range between  four and five millimeters, which might be even smaller in the case of cervical stenosis \cite{BETTOCCHI20091}. The level of its flexion varies between 115 and 185 degrees \cite{CagnacciAngelo2014Iomp, ColMadendagIlknur2020Eota}. Moreover, the cervical tissue is highly fragile and sensitive. Thus, any abrupt motion can cause pain to the patient, tissue damage, or even a uterine perforation. Based on the fact that a hysteroscope has similar diameter to the canal, introducing the instrument into the uterine cavity is never trivial for a clinician. In order to be able to perform the procedure in a safe and efficient manner, novice gynecologists should learn how to perform this maneuver in a way that does not harm their patients.

Our prior work has presented a platform for hysteroscopic training with the focus on the passage of the cervical canal \cite{PoliakovVladimir2020AVRS}\cite{Poliakov22}. It demonstrated the potential of this platform in medical education and training. To improve the quality of training and establish a more efficient assessment tool, we are aiming to design the virtual instructor feature that will provide a trainee with online cues on the optimal motion in the current state. In addition, it can also be used to assess the effectiveness and accuracy of a trainee's motion at each time step, thus allowing him/her to better analyze performance and identify wrong manipulations. With this work, we present the first step in fulfilling this goal, which is the design and evaluation of a \gls{rl} algorithm for automatic navigation of the hysteroscope in the task of the passage of the cervical canal. The developed algorithm is based on the \gls{sac} approach to smooth out the reward function during the learning process and ensure uniform exploration of the workspace. The developed algorithm was trained using two datasets: one acquired from expert clinicians performing the same task on the simulator and the other generated by the agent during the learning process. The agent's performance was then compared to the expert's performance using multi-metric approach.

\section{RELEVANT WORK}

A broad body of research has been dedicated to autonomous \gls{rl} path planning and navigation in the surgical simulation domain. Since the goal of the designed system was to provide stable results in any given state, one of the main topics of interest for us was how different approaches encourage the explorational behaviour of an agent. Some of the reviewed studies addressed this issue by injecting additional noise in the output of the agent's network. Nguyen et al. designed a system to automate the tensioning in the task of surgical pattern cutting \cite{Nguyen19a, Nguyen19b}. The underlying \gls{trpo} agent aims to locate the optimal pinch point at each segment of the pattern, then use forceps to grasp the tissue and control the applied force to maintain the optimal level of tension for cutting.
Similarly, Wenqiang et al. presented a system for automatic endovascular catheterization \cite{Wenqiang18}. The authors used the \gls{pi2} approach to train an agent that performs catheterization avoiding unwanted contacts between the catheter tip and the vessel wall. Both of the mentioned works control the exploitation/exploration balance by controlling the level of noise in the final output. This approach, in our opinion, underuses the exploration behaviour of the agent. While it might be sufficient for the task of path planning from a starting point, in our case, it might lead to unpredictable behaviour in dynamic path planning for the states that were not explored during the training phase.

Another group of the reviewed studies is based on the entropy-regularized \gls{rl}. Xiaoyu et al. presented a framework for robot-assisted  surgery training using deep \gls{rl} \cite{Xiaoyu19}. In their work, the authors trained a \gls{ppo} agent to perform the control policy in a simulation of the peg-and-hole exercise. Subsequently, an imitation agent was trained by fusing the learned policy and the  trajectories of experts who performed the same task in the real environment. The resulting system aims to provide trainees with demonstrated trajectories and feedback scores during practice. The agent's algorithm used in the paper, \gls{ppo}, was augmented with an entropy term in the reward function to improve the explorational behaviour of the agent, which eliminates aforementioned problem. On the other hand, the fact that this algorithm employs online training does not allow it to be used for off-policy learning, including learning by example, which necessitates additional steps to combine the behaviour of an agent with the behaviour of an expert.

An algorithm that features both off-policy training and entropy regularization is the \gls{sac} method. The key difference of this approach is that it controls the explorational behaviour of an agent by introducing the entropy term in the reward function. The aim of the agent is then to maximize both the reward and the entropy.
This can theoretically lead to learning a policy with a higher reward score and a smoother learning curve. 
As an example, Prianto et al. designed an algorithm for path planning for multi-arm manipulators using \gls{sac} \cite{Prianto20} . The performance of the agent was compared against efficiency of TD3 and \gls{sac} with no entropy. The article demonstrated faster learning rate and higher reward score of the \gls{sac} agents. Having slower learning rate, the \gls{sac} agent with entropy provided a higher reward score after approximately 100 episodes.
To accelerate the training process and increase the stability of the agent, the \gls{sac} approach can be combined with the \gls{her} method \cite{Andrychowicz18}. Thus, Gandana et al. designed a system for the reaching task of a scrub nurse \cite{Gandana20}. The system used a \gls{sac} agent together with the \gls{her}  method. The experiments demonstrated that the algorithm can deliver stable results even in the case of unstated spatial goal constraints.

\section{MATERIALS AND METHODS}

This section provides an overview of the designed training platform and the methods we used to build and train the agent. We first start with the description of the simulation, then proceed with the agent's underlying \gls{rl} approach, \gls{sac}. Finally, we define the reward mechanism that we used to train the agent and describe the training process itself.

\subsection {Simulation Platform}
\begin{figure}[ht]
    \centering
    \includegraphics[width=0.9\linewidth]{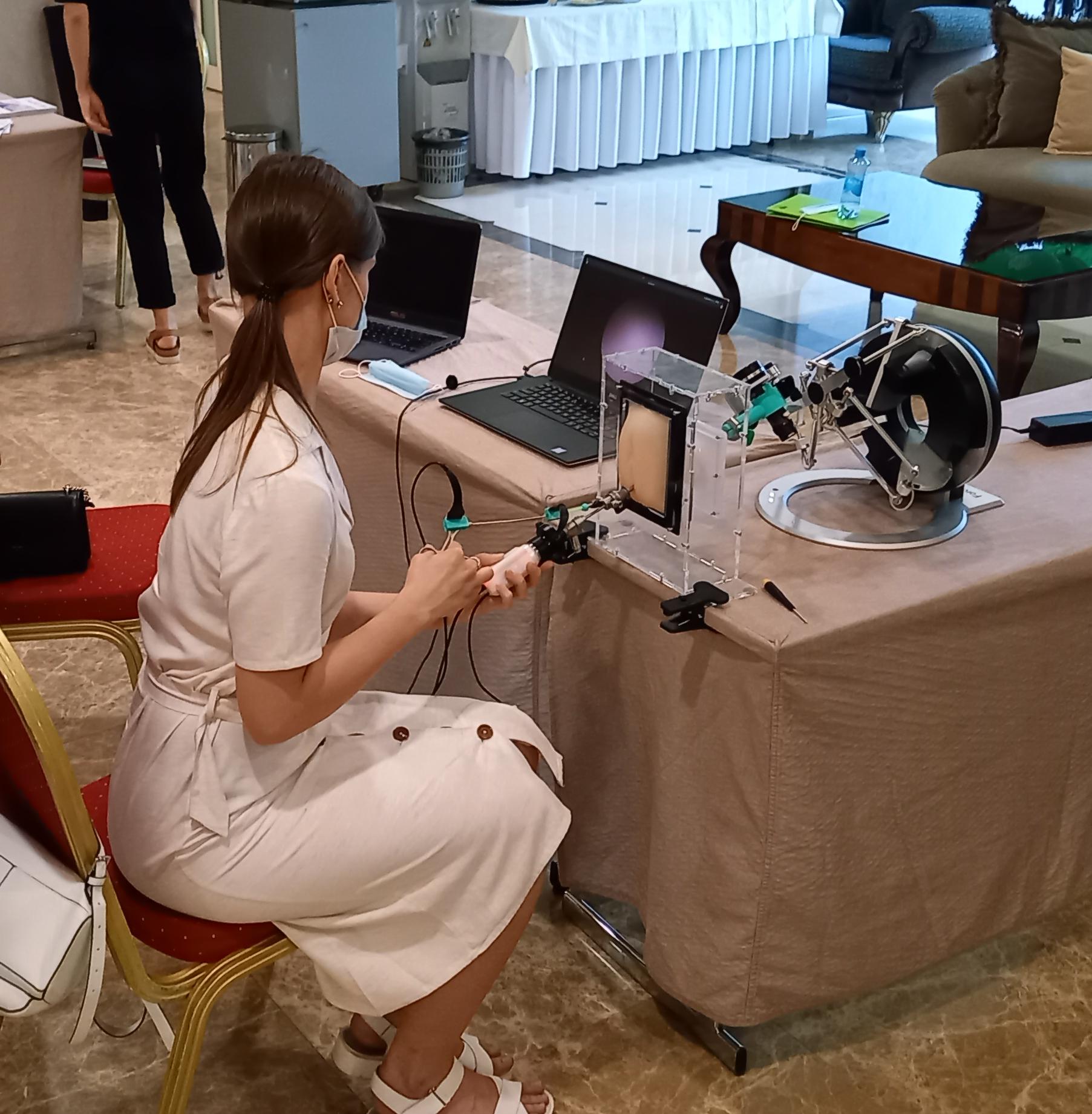}
    \caption{The view upon the simulation platform for in-office hysteroscopy.}
    \label{fig:exterior}
\end{figure}

\begin{figure}[ht]
    \centering
    \hfill
    \includegraphics[width=0.9\linewidth]{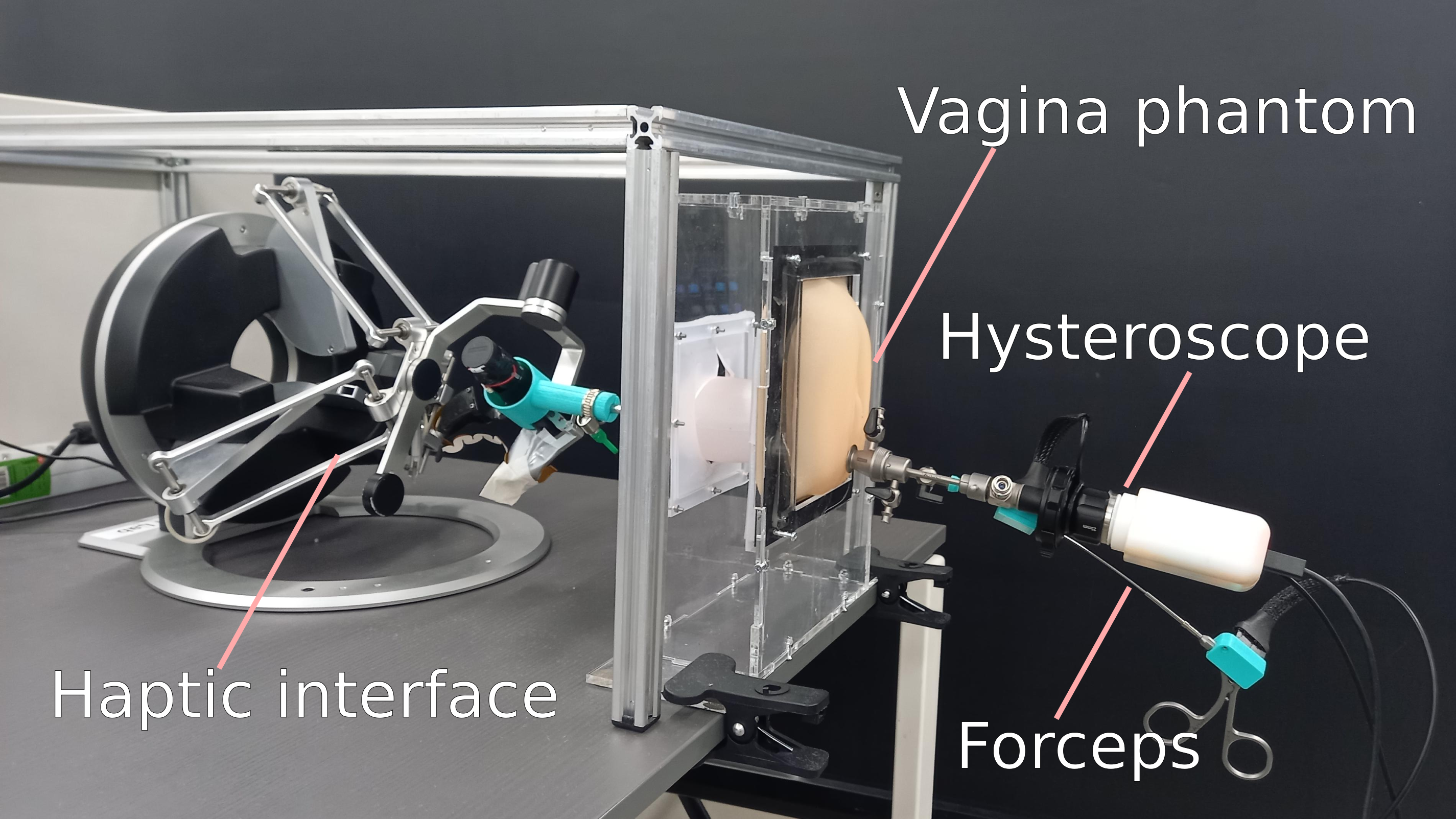}
    \caption{The hardware layout of the simulation platform for in-office hysteroscopy.}
    \label{fig:hw}
\end{figure}

The developed platform allows clinicians to practise essential skills for in-office hysteroscopy. The platform features an exercise for hand-eye coordination, camera navigation and basic instrumentation skills with the emphasis on the passage of the cervical canal. Figure \ref{fig:exterior} depicts the exterior view of the platform. A user operates the hysteroscope inserted through the phantom of the vagina (figure \ref{fig:hw}). The distal part of the hysteroscope is attached to the omega.7 haptic interface (Force Dimensions, Switzerland), which is used to track the position of the hysteroscope and to bind it to its virtual avatar, as well as to provide force feedback to the user.

From the software point of view, the simulation is comprised of the following entities: the uterus, the hysteroscope, and the checkpoints. To facilitate realistic soft-body deformation while maintaining the necessary update rate, the uterus is represented by three synchronized models, each performing a dedicated task. The visual model is used for graphical representation of the organ on the screen; it is a high quality mesh of 28910 faces. The mechanical model is used to simulate deformations of the body based on the \gls{fem}; it is a volumetric mesh consisting of 2873 tetrahedral elements. Finally, the collision model is used to detect intersections between the body and other elements in the scene. The collision model is a surface mesh of 3000 triangles.

In order to generate the uterus model, we used the Visible Human cryosection dataset \cite{Ackerman1998}. Using manual segmentation and MeshLab \cite{Meshlab} remeshing filters, we generated the visual model, which was then simplified to generate the collision and the volumetric meshes (the latter was created using CGAL and Delauney decomposition \cite{Alliez2016}).

The simulation relies on two main components running in the asynchronous mode. The physics engine is based on the SOFA framework and runs at approximately 150 \gls{fps}. This component performs collision detection and calculation of collision response and soft-body deformation. The second component, the visual loop, renders the visual models of the simulated entities onto the screen and runs at the frequency of 50 \gls{fps}. To bind the visual model of the uterus to its mechanical representation, we used barycentric mapping \cite{Floater2010}.

\subsection{Reinforcement Learning}
\label{sec:reinforcement_learning}

As stated above, the primary goal of this work is to design an agent that is capable of performing automatic introduction of the instrument into the uterine cavity through the cervical canal. The main requirement for the algorithm is to perform this task as fast as possible, while maintaining an adequate force profile to prevent any damage to the cervical tissue. To achieve this task, the \gls{sac} method was used \cite{Haarnoja18}. This algorithm ensures even exploration of the workspace along with a smooth learning curve. The following sections provide an overview of the algorithm,  the description of the employed reward model and the learning process.

\subsubsection{Soft Actor Critic}

The \gls{sac} method was first presented by Haarnoja et al. in 2018. This is an off-policy actor-critic deep reinforcement learning algorithm based on the maximum entropy \gls{rl} framework. The actor aims to maximize the expected reward while also maximizing entropy. A typical \gls{rl} algorithm seeks to maximize the expected sum of rewards:

\begin{equation}
    J(\pi) = \sum_{t=0}^T\mathbb{E}_{(\textbf{s}_t,\textbf{a}_t) \thicksim \rho_{\pi}}[r(\textbf{s}_t, \textbf{a}_t)],
\end{equation}

where $\pi$ is the policy of the agent. $\textbf{s}_t$ and $\textbf{a}_t$ are the state-action pair at time $t$, and $r$ is the reward received after the transition at time $t$.

In contrast to that, the maximum entropy objective encourages exploratory behaviour of a stochastic agent by augmenting the reward function with the entropy component:

\begin{equation}
    J(\pi) = \sum_{t=0}^T\mathbb{E}_{(\textbf{s}_t,\textbf{a}_t) \thicksim \rho_{\pi}}[r(\textbf{s}_t, \textbf{a}_t) + \alpha \mathcal{H}(\pi(\cdot | \textbf{s}_t))] ,
\end{equation}

where $\mathcal{H}$ is the entropy of the policy and $\alpha$ is the temperature coefficient defining the importance of the entropy against the reward.

The soft actor critic method is composed by three components: the value function $V_\psi(\textbf{s}_t)$ estimator, the soft critic estimating the Q-value function $Q_\theta(\textbf{a}_t, \textbf{s}_t)$, and the actor performing estimation of the optimal policy $\pi_\phi(\textbf{a}_t | \textbf{s}_t)$. Each of the components is implemented as a deep neural network with the parameters $\psi$, $\theta$, and $\phi$, respectively. The soft value function is trained to minimize the squared residual error:

\begin{equation}
    J_V(\psi) = \mathbb{E}_{\textbf{s}_t \thicksim \mathcal{D}}[\frac{1}{2} (V_\psi (\textbf{s}_t) - \mathbb{E}_{\textbf{a}_{t} \thicksim \pi_\phi}
    [Q_\theta (\textbf{s}_t, \textbf{a}_t) - log_{\pi_\phi}(\textbf{a}_t | \textbf{s}_t)])^2],
\end{equation}

where $\mathcal{D}$ is the replay buffer, i.e. the distribution of previously sampled states. The soft Q-function is trained to minimize the soft Bellman residual:

\begin{equation}
    J_Q (\theta) = \mathbb{E}_{(\textbf{s}_t, \textbf{a}_t) \thicksim \mathcal{D})}[\frac{1}{2} (Q_\theta(\textbf{s}_t, \textbf{a}_t) - \hat{Q}_\theta(\textbf{s}_t, \textbf{a}_t))^2],
\end{equation}

with

\begin{equation}
    \hat{Q}(\textbf{s}_t, \textbf{a}_t) = r (\textbf{s}_t, \textbf{a}_t) + \gamma \mathbb{E}_{\textbf{s}_{t+1} \thicksim p}[V_{\bar{\psi}} (\textbf{s}_{t+1})]
\end{equation}

Finally, the policy can be trained by minimizing the Kullback-Leibler divergence:

\begin{equation}
    J_\pi (\phi) = \mathbb{E}_{\textbf{s}_t \thicksim \mathcal{D}} \left[\kldiv*{\pi_\phi (\cdot | \textbf{s}_t)}{\frac{\exp(Q_\theta (\textbf{s}_t, \cdot))}{Z_\theta (\textbf{s}_t)}} \right],
\end{equation}

where $Z_\theta (\textbf{s}_t)$ is the partition function that normalizes the distribution and, in principle, can be ignored \cite{Haarnoja18} .

\begin{figure}[ht]
    \centering
    \includegraphics[width=0.8\linewidth]{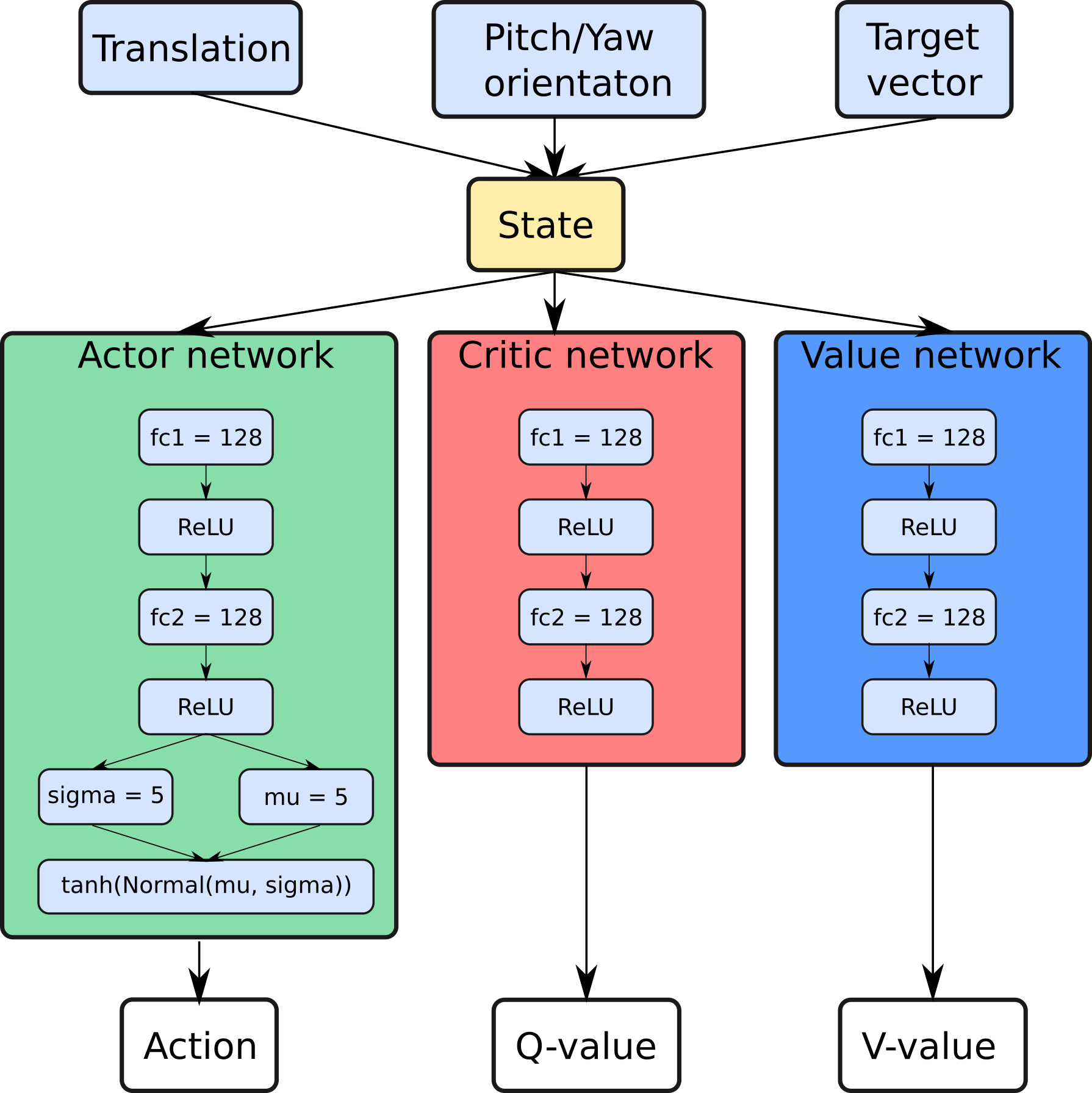}
    \caption{Input, output, and internal network parameters of the designed \gls{sac} agent. For each layer, the number of neurons is specified. The abbreviation \textit{fc} stands for fully connected layer.}
    \label{fig:networks}
\end{figure}

Figure \ref{fig:networks} depicts the neural network parameters we used in the developed algorithm. Each of the networks contains two hidden layers with 128 nodes and the input layers of the state's size. The state $\textbf{s}_t$ consists of four elements: position of the instrument, orientation of the instrument (given in Euler's angles), target vector (direction of the next checkpoint), and force feedback. The actor picks the next action based on these four parameters. The actor network generates mean value $\mu$ and standard deviation $\sigma$ to generate a normal distribution for five variables of two parameters: three for translation and two for rotation in pitch and yaw. Each increment is then randomly picked using generated $\mu$ and $\sigma$ to implement stochastic behaviour.

\subsubsection{Reward Model}

The proposed reward model combines dense and sparse rewards to improve the stability of training. Reward $r(\textbf{s}_t, \textbf{a}_t)$ is defined by the following equation:

\begin{equation}
    r(\mathbf{s}_t, \mathbf{a}_t) = r_c  - r_F - r_t - r_d,
\end{equation}

where $r_c$ is a sparse reward received every time the agent reaches a checkpoint; $r_F = F \cdot k_F$ is the reward associated with the applied force $F$;
$r_t = dt \cdot k_t$ is the reward associated with the time elapsed between transitions $dt$, and $r_d = d_c \cdot k_d$ is the reward associated with the distance to the next checkpoint $d_c$. In principle, the reward reflects the task imposed on the agent. It should perform the exercise in the minimal time frame while maintaining an adequate force level. Both time and force rewards are negative, which means that the agent will try to perform the task faster to minimize the penalty. But it also means that it can learn a false policy and instead avoid contact with the uterus to reduce the size of the force reward $r_F$. Thus, to stimulate the expected behaviour, we added two rewards in the equation. The first component $r_c$ is a positive sparse reward that is given every time the agent reaches one of the checkpoints located along the path inside of the cervical canal. 
The second component $r_d$ is a negative dense reward that is proportional to the distance between the hysteroscope and the next checkpoint. The idea behind adding this component is to speed up the learning process by drawing a direct link between the target vector provided in the state space and the size of the reward. 

\subsubsection {Learning Process}

\begin{figure}[ht]
    \centering
    \includegraphics[width=\linewidth]{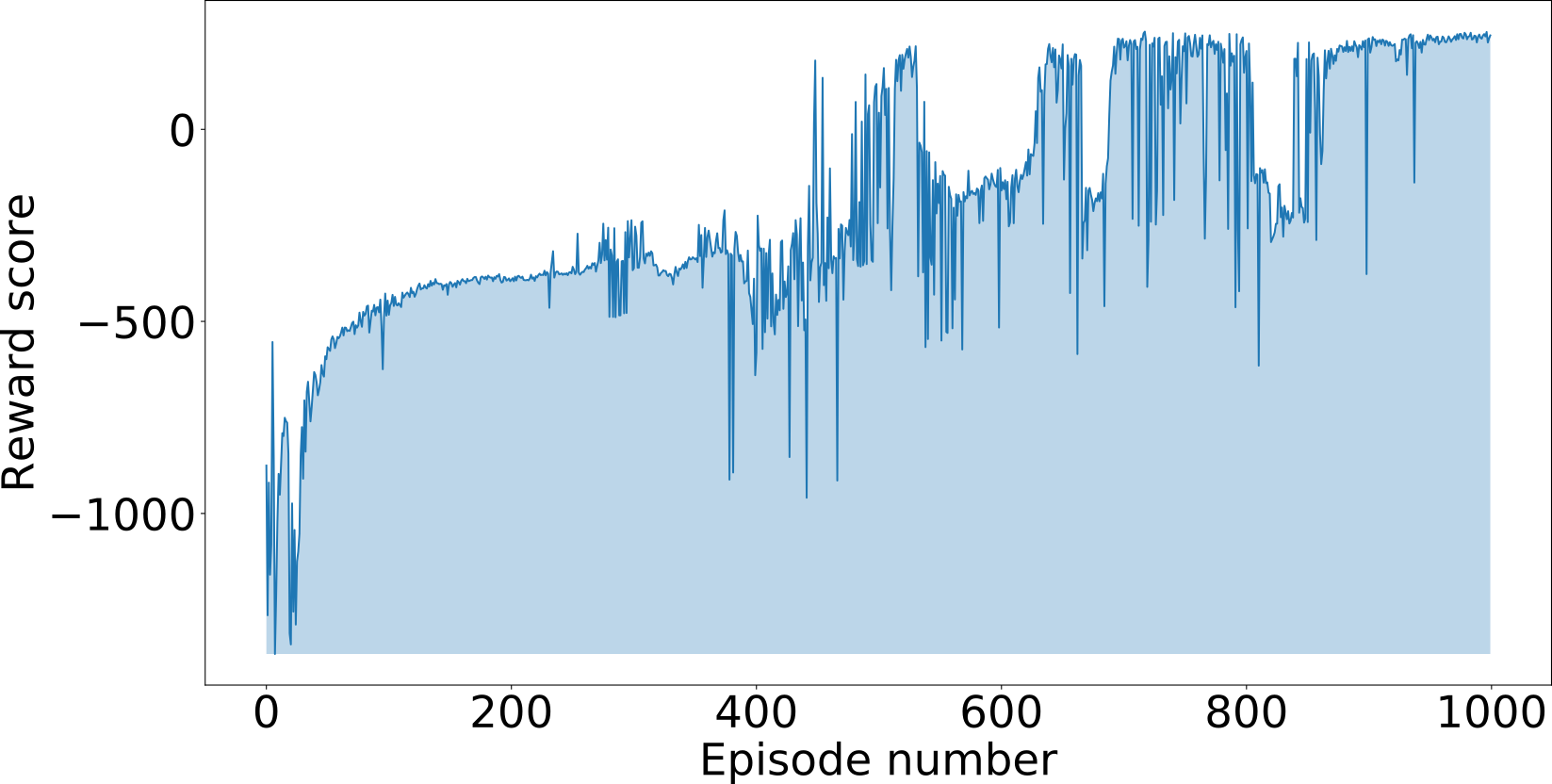}
    \caption{The plot showing the acquired rewards during the training of the \gls{sac} agent.}
    \label{fig:reward}
\end{figure}

The offline nature of the algorithm allowed us to employ both on-policy and off-policy training to accelerate the learning process. The learning process included two phases. First, the initial training of the agent was performed using the data acquired from expert clinicians. After 1000 updates we started the simulation to acquire the initial replay buffer. Once the minimally required batch was recorded, the agent started updating the networks after each five steps. The agent updated the networks using the following pattern: one update using the the data from expert clinicians after four updates using the replay buffer. The maximum time for an episode was set to 20 seconds. The number of episodes was 1000. Figure \ref{fig:reward} depicts the reward graph of this phase.

One of the greater challenges of implementing this approach is associated with the learning process. Due to the fact that we have two opposite constraints expressed in the reward components $r_F$ and $r_t$,
the agent becomes sensitive to the chosen coefficients for these components and does not always converge to the optimal or sub-optimal solution, rather
learning to stay idle in the proximity of one of the checkpoints.
Following values were chosen in the final design to successfully train the agent: force reward constant $k_F$ was set to 2.0, time reward constant $k_t$ was set to 0.1, sparse checkpoint reward $r_c$ was set to 200, and distance constant $k_d$ was set to 0.4.

\section{EXPERIMENTS}

To test the proposed method, we used the data acquired from expert clinicians and compared their performance with the performance of the developed \gls{sac} agent. Clinicians operated the hysteroscope attached to the haptic model of the female reproductive system as shown in figure \ref{fig:exterior} to perform the exercise. For clinicians, the task was the same as for the agent: to introduce the hysteroscope into the uterine cavity. Clinicians were also asked to perform the procedure in a regular way, meaning applying the same force as they would usually do in a real procedure. Based on the acquired data, we calculated the following metrics:

\begin{enumerate}
    \item \textit{Maximal force $F_{max}$}: modulus of the maximal applied force.
    \item \textit{Integral force $F_{i}$}: integral value of the modulus of applied force.
    \item{Force \gls{fft} $F_{FFT}$}: integral value of force amplitude in the frequency domain in the range between $1$ and $13$~Hz corresponding to the frequency range of voluntary motion and physiological tremor.
    \item \textit{Execution time $t_e$}: time to successfully finish the exercise.
\end{enumerate}

We already used some of the metrics in our prior work \cite{PoliakovVladimir2020AVRS} \cite{Poliakov22}, in which they demonstrated the ability to differentiate between novices and experts. Integral force $F_i$ is a cross-metric affected by both the modulus of force and time. That is why we reckon this metric is an adequate way to assess the level of patient comfort.
Force \gls{fft} $F_{FFT}$ is the metric showing tremor and oscillation in the voluntary motion, which can reflect the level of confidence and dexterity. Combined, these four metrics can provide a reasonable overview of differences in the behaviour of the agent and the expert clinicians.

\section{RESULTS AND DISCUSSION}

\begin{figure}[ht]
    \centering
    \includegraphics[width=0.8\linewidth]{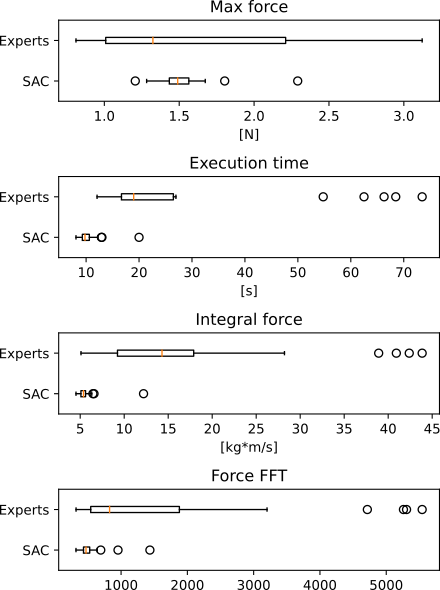}
    \caption{The box plots demonstrating median and interquartile values of the recorded metrics.}
    \label{fig:results}
\end{figure}

The performance of the \gls{sac} agent was analysed based on 50 episodes and compared with the performance of five experts, each of which performed the exercise five times. Out of 50 attempts, 49 cases were completed successfully. The results of the experiments are presented in Figure \ref{fig:results}. Median value of maximal force $F_{max}$ for the \gls{sac} agent was $1.49N \pm0.15 SD$ against $1.32 N \pm0.79SD$ for the expert group.
Median value of execution time $t_e$ was $9.79 s \pm 1.79SD$ for the agent and $18.98 s \pm 19.31SD$ for the expert group. Median value of integral force was $5.37 kg \cdot m / s \pm 1.05SD$ for the \gls{sac} agent against $14.29 kg \cdot m / s \pm 11.59SD$ for the expert group. Median value of force \gls{fft} $F_{FFT}$ was $471.11 \pm 168.22SD$ for the agent versus $826.84 \pm 1672.70SD$ for the expert group.

\begin{figure}[ht]
    \centering
    \includegraphics[width=0.8\linewidth]{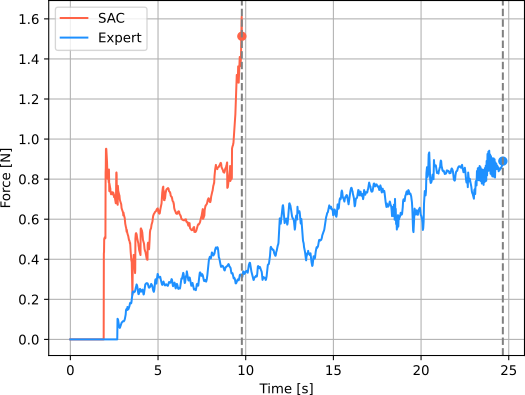}
    \caption{The plot showing the force profile of randomly sampled attempts of the agent and an expert.}
    \label{fig:force_profile}
\end{figure}

Overall, the agent demonstrated satisfactory results and outperformed the expert group in three metrics out of four. Although median value of maximal force was higher than that of the expert group, all agent's recorded attempts still fall within the interquartile range of the experts' recorded attempts.
For integral force $F_{i}$ and and force FFT ($F_{FFT}$), the results of the agent were significantly better compared to the experts' performance. This might be related to two factors. First, the agent performed the exercise in a shorter time frame with a comparable level of applied force. Figure \ref{fig:force_profile} depicts the force profile of randomly sampled attempts of the agent and an expert. Second, the stochastic nature of the agent's behaviour decreased over the course of training leading to a more deterministic behaviour as the actor network was being optimised (figure \ref{fig:networks}). This, in turn, explains smaller oscillations in motion. 

It is also worth mentioning that although the agent demonstrated exquisite performance, it cannot yet be generalized and applied to any given anatomy of the uterus. While the agent was able to outperform the experts after 1000 episodes, it took him more than 500 episodes to simply learn how to complete the task, let alone performing it in an optimal manner. In contrast, the experts were able to complete all five attempts without additional training and the results remained consistent throughout the experiments, although with a wider spread. Thus, as the agent was optimised for a particular scene, additional work will be required to generalise this approach for a broader range possible anatomies.

\section{CONCLUSION}

In this work, we presented an \gls{rl}-based algorithm for automatic passage of the cervical canal in the task of outpatient hysteroscopy simulation. The concept was tested with expert clinicians to prove the feasibility of the proposed approach. The agent performed better compared to the expert group in three out of four recorded metrics. The fourth metric, maximal force, was also within the interquartile range of the experts' results. Overall, the algorithm demonstrated high efficiency and reliability.

Future work will focus on the implementation of an online assistance tool for outpatient hysteroscopy training based on the designed agent. The task of this assistant will be to guide the trainees by providing online cues about the direction of motion and the level of applied force.  Once designed, the system will be compared with conventional training techniques with an instructor.

Another possible point of improvement is to change the type of the state data that is fed to the agent. While a user selects actions based on the image data on the screen and kinaesthetic information (force feedback and the pose of the hand), he/she has no information about the position of the checkpoints relatively to the current position of the hysteroscope. This means that the agent makes decisions using the information not available to a person, which is why, at this moment, it cannot fully mimic the behaviour of an expert for different configurations of the uterus. To overcome this challenge, we will design and train the agent with the state comprised only of those pieces of information that are also available to a user:
pose of the hysteroscope, image data and force feedback. Using this approach can lead to a more general solution that works for different anatomical structures.

Overall, the developed \gls{rl}-based algorithm for autonomous surgical instrument introduction for outpatient hysteroscopy training creates a promising opportunity to be integrated in the haptic training platform to offer online cues for educational instruction and data-driven based assessment tools for evaluating the performance of a trainee. The presented work certainly facilitate the integration of \gls{rl}-based methods with haptics and robotic system control.

\section{ACKNOWLEDGEMENTS}

The reported study was funded by RFBR and CNRS
according to the research project No. 21-58-15006.

\addtolength{\textheight}{-12cm}   



\bibliographystyle{IEEEtran}
\bibliography{IEEEabrv,bibliography.bib}

\begin{thebibliography}{10}
\providecommand{\url}[1]{#1}
\csname url@samestyle\endcsname
\providecommand{\newblock}{\relax}
\providecommand{\bibinfo}[2]{#2}
\providecommand{\BIBentrySTDinterwordspacing}{\spaceskip=0pt\relax}
\providecommand{\BIBentryALTinterwordstretchfactor}{4}
\providecommand{\BIBentryALTinterwordspacing}{\spaceskip=\fontdimen2\font plus
\BIBentryALTinterwordstretchfactor\fontdimen3\font minus
  \fontdimen4\font\relax}
\providecommand{\BIBforeignlanguage}[2]{{%
\expandafter\ifx\csname l@#1\endcsname\relax
\typeout{** WARNING: IEEEtran.bst: No hyphenation pattern has been}%
\typeout{** loaded for the language `#1'. Using the pattern for}%
\typeout{** the default language instead.}%
\else
\language=\csname l@#1\endcsname
\fi
#2}}
\providecommand{\BIBdecl}{\relax}
\BIBdecl

\bibitem{BETTOCCHI20091}
\BIBentryALTinterwordspacing
S.~Bettocchi, A.~{Di Spiezio Sardo}, and O.~Ceci, ``Chapter 1 - instrumentation
  in office hysteroscopy: Rigid hysteroscopy,'' in \emph{Hysteroscopy}, L.~D.
  Bradley and T.~Falcone, Eds.\hskip 1em plus 0.5em minus 0.4em\relax
  Philadelphia: Mosby, 2009, pp. 1--6. [Online]. Available:
  \url{https://www.sciencedirect.com/science/article/pii/B9780323041010500072}
\BIBentrySTDinterwordspacing

\bibitem{vanWesselSteffi2018HitN}
S.~van Wessel, T.~Hamerlynck, B.~Schoot, and S.~Weyers,
  ``\BIBforeignlanguage{eng}{Hysteroscopy in the netherlands and flanders: A
  survey amongst practicing gynaecologists},''
  \emph{\BIBforeignlanguage{eng}{European journal of obstetrics \& gynecology
  and reproductive biology}}, vol. 223, pp. 85--92, 2018.

\bibitem{Hernandez2018}
A.~U. Hernandez, \emph{In-Office Hysteroscopy}.\hskip 1em plus 0.5em minus
  0.4em\relax Cham: Springer International Publishing, 2018, pp. 33--40.

\bibitem{CagnacciAngelo2014Iomp}
A.~Cagnacci, G.~Grandi, M.~Cannoletta, A.~Xholli, I.~Piacenti, and A.~Volpe,
  ``\BIBforeignlanguage{eng ; fre ; ger}{Intensity of menstrual pain and
  estimated angle of uterine flexion},'' \emph{\BIBforeignlanguage{eng ; fre ;
  ger}{Acta obstetricia et gynecologica Scandinavica}}, vol.~93, no.~1, pp.
  58--63, 2014.

\bibitem{ColMadendagIlknur2020Eota}
I.~Col~Madendag, M.~Eraslan~Sahin, Y.~Madendag, E.~Sahin, M.~B. Demir,
  F.~Ozdemir, G.~Acmaz, and I.~I. Muderris, ``\BIBforeignlanguage{eng}{Effect
  of the anterior uterocervical angle in unexplained infertility: a prospective
  cohort study},'' \emph{\BIBforeignlanguage{eng}{Journal of international
  medical research}}, vol.~48, no.~4, pp. 497--507, 2020.

\bibitem{PoliakovVladimir2020AVRS}
V.~Poliakov, B.~De~Vree, D.~Tsetserukou, K.~Niu, and E.~Vander~Poorten, ``A
  virtual reality surgical training system for office hysteroscopy with haptic
  feedback: A feasibility study,'' in \emph{Virtual Reality and Augmented
  Reality}.\hskip 1em plus 0.5em minus 0.4em\relax Springer, Cham, 2020, pp.
  115--127.

\bibitem{Poliakov22}
V.~Poliakov, K.~Niu, D.~Tsetserukou, and E.~V. Poorten, ``An in-office
  hysteroscopy vr/haptic simulation platform for training in spatial navigation
  and passage of the cervical canal,'' \emph{IEEE Transactions on Medical
  Robotics and Bionics}, vol.~4, no.~3, pp. 608--620, 2022.

\bibitem{Nguyen19a}
T.~Nguyen, N.~D. Nguyen, F.~Bello, and S.~Nahavandi, ``A new tensioning method
  using deep reinforcement learning for surgical pattern cutting,'' in
  \emph{2019 IEEE International Conference on Industrial Technology (ICIT)},
  2019, pp. 1339--1344.

\bibitem{Nguyen19b}
N.~D. Nguyen, T.~Nguyen, S.~Nahavandi, A.~Bhatti, and G.~Guest, ``Manipulating
  soft tissues by deep reinforcement learning for autonomous robotic surgery,''
  in \emph{2019 IEEE International Systems Conference (SysCon)}, 2019, pp.
  1--7.

\bibitem{Wenqiang18}
W.~Chi, J.~Liu, M.~E. M.~K. Abdelaziz, G.~Dagnino, C.~Riga, C.~Bicknell, and
  G.-Z. Yang, ``Trajectory optimization of robot-assisted endovascular
  catheterization with reinforcement learning,'' in \emph{2018 IEEE/RSJ
  International Conference on Intelligent Robots and Systems (IROS)}, 2018, pp.
  3875--3881.

\bibitem{Xiaoyu19}
X.~Tan, C.-B. Chng, Y.~Su, K.-B. Lim, and C.-K. Chui, ``Robot-assisted training
  in laparoscopy using deep reinforcement learning,'' \emph{IEEE Robotics and
  Automation Letters}, vol.~4, no.~2, pp. 485--492, 2019.

\bibitem{Prianto20}
\BIBentryALTinterwordspacing
E.~Prianto, M.~Kim, J.-H. Park, J.-H. Bae, and J.-S. Kim, ``Path planning for
  multi-arm manipulators using deep reinforcement learning: Soft actor–critic
  with hindsight experience replay,'' \emph{Sensors}, vol.~20, no.~20, 2020.
  [Online]. Available: \url{https://www.mdpi.com/1424-8220/20/20/5911}
\BIBentrySTDinterwordspacing

\bibitem{Andrychowicz18}
M.~Andrychowicz, F.~Wolski, A.~Ray, J.~Schneider, R.~Fong, P.~Welinder,
  B.~McGrew, J.~Tobin, P.~Abbeel, and W.~Zaremba, ``Hindsight experience
  replay,'' 2018.

\bibitem{Gandana20}
C.~E. Gandana, J.~D.~K. Disu, H.~Xie, and L.~Gu, ``Analyzing different unstated
  goal constraints on reinforcement learning algorithm for reacher task in the
  robotic scrub nurse application,'' in \emph{2020 IEEE International
  Conference on Industry 4.0, Artificial Intelligence, and Communications
  Technology (IAICT)}, 2020, pp. 42--47.

\bibitem{Ackerman1998}
M.~J. Ackerman, ``{The visible human project},'' \emph{Proceedings of the
  IEEE}, vol.~86, no.~3, pp. 504--511, 1998.

\bibitem{Meshlab}
P.~Cignoni, M.~Callieri, M.~Corsini, M.~Dellepiane, F.~Ganovelli, and
  G.~Ranzuglia, ``{MeshLab: an Open-Source Mesh Processing Tool},'' in
  \emph{Eurographics Italian Chapter Conference}, V.~Scarano, R.~D. Chiara, and
  U.~Erra, Eds.\hskip 1em plus 0.5em minus 0.4em\relax The Eurographics
  Association, 2008.

\bibitem{Alliez2016}
P.~Alliez and A.~Fabri, ``{CGAL-The computational geometry algorithms
  library},'' in \emph{ACM SIGGRAPH 2016 Courses, SIGGRAPH 2016}.\hskip 1em
  plus 0.5em minus 0.4em\relax Association for Computing Machinery, Inc, 2016.

\bibitem{Floater2010}
\BIBentryALTinterwordspacing
M.~S. Floater and J.~Kosinka, ``{Barycentric interpolation and mappings on
  smooth convex domains},'' in \emph{Proceedings - 14th ACM Symposium on Solid
  and Physical Modeling, SPM'10}.\hskip 1em plus 0.5em minus 0.4em\relax New
  York, New York, USA: ACM Press, 2010, pp. 111--116. [Online]. Available:
  \url{http://portal.acm.org/citation.cfm?doid=1839778.1839794}
\BIBentrySTDinterwordspacing

\bibitem{Haarnoja18}
T.~Haarnoja, A.~Zhou, P.~Abbeel, and S.~Levine, ``Soft actor-critic: Off-policy
  maximum entropy deep reinforcement learning with a stochastic actor,'' in
  \emph{Proceedings of the 35th International Conference on Machine Learning},
  ser. Proceedings of Machine Learning Research, J.~Dy and A.~Krause, Eds.,
  vol.~80.\hskip 1em plus 0.5em minus 0.4em\relax PMLR, 10--15 Jul 2018, pp.
  1861--1870.

\end{thebibliography}

\end{document}